\documentclass[conference]{IEEEtran}
\IEEEoverridecommandlockouts
\usepackage{cite}
\usepackage{makecell} 
\usepackage{svg} 
\usepackage{siunitx}

\usepackage{stfloats}   
\usepackage{booktabs} 
\usepackage{pifont}
\usepackage{booktabs}
\usepackage{float} 
\usepackage{xcolor}
\usepackage{array}
\renewcommand{\arraystretch}{1.2}
\usepackage{subcaption} 
\usepackage{caption}
\usepackage{pgfplotstable}
\pgfplotsset{compat=1.18}
\usepackage[table,xcdraw]{xcolor}
\usepackage{hyperref}
\usepackage{amsmath,amssymb,amsfonts}
\usepackage{algorithm}
\usepackage{algpseudocode}
\usepackage{graphicx}
\usepackage{textcomp}
\usepackage{xcolor}
\usepackage{multirow}
\usepackage{placeins}
\usepackage{ragged2e}

\usepackage{graphicx}        
\usepackage[export]{adjustbox}
\def\BibTeX{{\rm B\kern-.05em{\sc i\kern-.025em b}\kern-.08em
    T\kern-.1667em\lower.7ex\hbox{E}\kern-.125emX}}
\begin{document}

\title{Adversarial Robustness Analysis of Cloud-Assisted Autonomous Driving Systems}
\author{
  \IEEEauthorblockN{Maher Al Islam and Amr S. El-Wakeel}
  \IEEEauthorblockA{%
    Lane Department of Computer Science and Electrical Engineering\\
    West Virginia University, WV, USA}
}

\maketitle
\begin{abstract}
Autonomous vehicles increasingly rely on deep learning–based perception and control, which impose substantial computational demands. Cloud-assisted architectures offload these functions to remote servers, enabling enhanced perception and coordinated decision-making through the Internet of Vehicles (IoV). However, this paradigm introduces cross-layer vulnerabilities, where adversarial manipulation of perception models and network impairments in the vehicle–cloud link can jointly undermine safety-critical autonomy. This paper presents a hardware-in-the-loop IoV testbed that integrates real-time perception, control, and communication to evaluate such vulnerabilities in cloud-assisted autonomous driving. A YOLOv8-based object detector deployed on the cloud is subjected to white-box adversarial attacks using the Fast Gradient Sign Method (FGSM) and Projected Gradient Descent (PGD), while network adversaries induce delay and packet loss in the vehicle–cloud loop. Results show that adversarial perturbations significantly degrade perception performance, with PGD reducing detection precision and recall from 0.73 and 0.68 in the clean baseline to 0.22 and 0.15 at $\epsilon = 0.04$. Network delays of 150–250\,ms, corresponding to transient losses of approximately 3–4 frames, and packet loss rates of 0.5–5\% further destabilize closed-loop control, leading to delayed actuation and rule violations. These findings highlight the need for cross-layer resilience in cloud-assisted autonomous driving systems.
\end{abstract}

\thispagestyle{empty}
\pagestyle{empty}


\section{Introduction}

Autonomous vehicles (AVs) are envisioned as the backbone of next-generation intelligent transportation systems, promising to reduce human error and enhance road safety. Yet, a series of real-world incidents, from Tesla’s perception failures to Uber’s fatal pedestrian crash, underscore persistent vulnerabilities in AV perception and decision-making pipelines~\cite{liu2024exploration}. Despite major advances in sensing, control, and connectivity, modern AVs remain susceptible to safety-critical disruptions under uncertain environmental and cyber-physical conditions~\cite{yang2023anomaly}. With nearly 1.9 million global road fatalities each year~\cite{who2024roadtraffic}, ensuring trustworthy and robust autonomy has become a high priority for intelligent transportation systems.

To evaluate and validate autonomous driving stacks, most research has relied on simulation platforms such as CARLA, MetaDrive, and ApolloScape~\cite{li2022metadrive}, or benchmark datasets like KITTI, nuScenes, and Cityscapes~\cite{apruzzese2023real}, among others. While these resources support scalable experimentation, they inherently abstract away real-world uncertainties such as sensor degradation, stochastic disturbances, and network latency. In contrast, physical testbeds such as Duckietown~\cite{paull2017duckietown} enable closed-loop, real-time experimentation of perception and control modules under degraded sensing and communication. By bridging simulation-based assurance and real-world operation, such testbeds offer a pragmatic foundation for studying system-level robustness in IoT-enabled autonomy~\cite{mokhtarian2024survey}.

Building on this foundation, recent research in AV has shifted toward cloud-assisted architectures that distribute perception and control intelligence between vehicles and cloud servers. In these systems, on-vehicle clients perform sensing and actuation only, while the cloud executes computationally intensive modules for perception, prediction, and decision-making~\cite{xu2023cloud, schafhalter2023leveraging, zhao2025distributed}. For example, Schafhalter et al. introduced a hierarchical task allocation framework that dynamically partitions workloads between the edge and cloud to balance latency and inference accuracy~\cite{schafhalter2023leveraging}. Zhao et al. extended this concept through distributed coordination, enabling multiple vehicles to share intermediate representations via the cloud for cooperative perception and control~\cite{zhao2025distributed}. Collectively, these frameworks demonstrate the operational advantages of centralized intelligence, but also expose new cross-layer vulnerabilities, where perception degradation and network impairment can cascade through the control loop and compromise driving safety.


Adversarial AI exploits the vulnerability of deep neural networks where small human-imperceptible perturbations can drastically alter model predictions~\cite{oprea2023adversarial}. Such attacks are generally classified as \textit{white-box} or \textit{black-box}, depending on the attacker’s knowledge of the model’s architecture and parameters. In the white-box setting, the adversary has full access to the model’s parameters and gradients, enabling direct manipulation of detector outputs through methods such as Fast Gradient Sign Method (FGSM)~\cite{goodfellow2014explaining} and  Projected Gradient Descent (PGD)~\cite{madry2017towards}. In contrast, black-box attacks operate without internal access, relying on query-based or transfer-based strategies, including physical perturbations such as adversarial patches or reflective overlays~\cite{amirkhani2023survey}, to mislead models like YOLO~\cite{YOLOv8_ultralytics} and cause missed detections of critical objects such as stop signs and pedestrians~\cite{akhtar2021advances}.

Concurrently, network-layer adversaries disrupt cloud communication by injecting delay, packet loss, or replayed data, destabilizing control synchronization between vehicles and the cloud. Previous studies have shown that even modest latency or loss can propagate timing errors through the control stack, resulting in unsafe maneuvers or rule violations~\cite{giannaros2023autonomous, sun2021survey, gupta2023investigation}. Network impairments intensify the impact of adversarial perception by disrupting the timely delivery and execution of vision-based control commands. While prior studies examine perception or communication robustness independently, few works experimentally quantify their joint impact in a real-world IoV setting.

To address these cross-layer threats, this paper presents an IoV testbed to assess the robustness of cloud-assisted autonomous driving systems under coordinated adversarial conditions. The framework integrates real-time perception, control, and communication modules, enabling systematic experimentation with adversarial AI and network-layer adversary. A white-box adversary perturbs visual inputs to degrade YOLOv8-based object detection, while a network adversary, modeled following the MITRE ATT\&CK framework~\cite{mitre_attack}, injects latency and packet loss in the vehicle–cloud loop. Together, these experiments provide a reproducible foundation for analyzing how perception degradation and network impairment compromise decision reliability and driving safety. The main contributions of this paper are summarized as follows:

\begin{itemize}
\item Development of a cloud-assisted IoV testbed integrating perception, control, and communication layers for the evaluation of adversarial robustness.
\item Implementation of white-box adversarial attacks (FGSM and PGD) on YOLOv8 object detection to assess robustness under adversarial perturbations.
\item Modeling of network-layer adversaries using the MITRE ATT\&CK framework to inject delay and packet loss effects on closed-loop control stability.
\item Evaluation of adversarial AI and network adversary to characterize their impact on perception accuracy and decision reliability in cloud-assisted autonomous driving.
\end{itemize}

The remainder of this paper is organized as follows. Section~II presents the technical background. Section~III introduces the IoV testbed architecture and threat models. Section~IV details the experimental setup for robustness evaluation, and Section~V discusses results across perception and control layers. Section~VI concludes with key insights on cross-layer robustness in cloud-assisted autonomous driving, and Section~VII acknowledges the support received for this work.

\section{Preliminaries}
This section provides an overview of the underlying models and adversarial methods that support our experimental framework. We outline the YOLOv8 detection backbone and its training loss, followed by gradient-based white-box adversarial methods such as FGSM and its iterative extension PGD, which are used for robustness evaluation.

\subsection{YOLO-based Object Detection}
YOLO is a single-stage object detection framework that jointly optimizes localization and classification to achieve real-time inference from raw visual inputs. The YOLOv8 variant builds upon this architecture with enhanced feature representation and detection efficiency, making it well suited for real-time perception in cloud-assisted IoV systems. Given an input image $x$, the network predicts bounding boxes $b_i$, class labels $c_i$, and confidence scores $p_i$ using composite loss:
\begin{equation}
\label{eq:yolov8_loss}
L = \lambda_{\text{box}} L_{\text{CIoU}} + \lambda_{\text{cls}} L_{\text{BCE}} + \lambda_{\text{dfl}} L_{\text{DFL}},
\end{equation}
where $L_{\text{CIoU}}$, $L_{\text{BCE}}$, and $L_{\text{DFL}}$ correspond to localization, classification, and distribution focal losses, respectively~\cite{YOLOv8_ultralytics}.

\subsection{Fast Gradient Sign Method (FGSM)}
FGSM is a single-step white-box attack that perturbs an input in the direction of the model loss gradient to produce an adversarial example. For a model with parameters \(\theta\), input \(x\) and true label \(y\), FGSM constructs
\begin{equation}
x' = x + \epsilon\,\mathrm{sign}\!\big(\nabla_x J(\theta,x,y)\big)
\end{equation}
where \(\nabla_x J(\theta,x,y)\) denotes the gradient of the model loss with respect to the input, guiding the perturbations toward directions that maximize prediction error. This is attributed to the near-linearity of deep networks in high-dimensional spaces, where many small, aligned perturbations can jointly cause large output deviations~\cite{goodfellow2014explaining}. As a single-step backpropagation-based attack under a constraint \(L_\infty\), FGSM efficiently generates worst-case samples and serves as a foundation for adversarial testing.

\subsection{Projected Gradient Descent (PGD)}
PGD extends FGSM by iteratively applying small gradient-based perturbations to maximize the model loss while constraining each update within a $\epsilon$-bounded $L_\infty$ ball around the original input~\cite{madry2017towards}. At each iteration, the perturbed sample is updated and projected back into the valid perturbation set.
\begin{equation}
x^{t+1} = \Pi_{x+\mathcal{S}}\big(x^{t} + \alpha\,\mathrm{sign}(\nabla_x J(\theta,x^{t},y))\big)
\end{equation}
where $\alpha$ is the step size, $\Pi$ denotes projection, and $\nabla_x J(\theta,x^{t},y)$ is the input gradient computed via backpropagation. This iterative formulation enables controlled exploration of the perturbation space and yields stronger more transferable attacks than single-step methods. Consequently, PGD is used as the primary mechanism for generating adversarial samples in experimental evaluation.

\section{System Architecture and Threat Model} 
\begin{figure*}[!t]
    \centering
    \includegraphics[width=\textwidth]{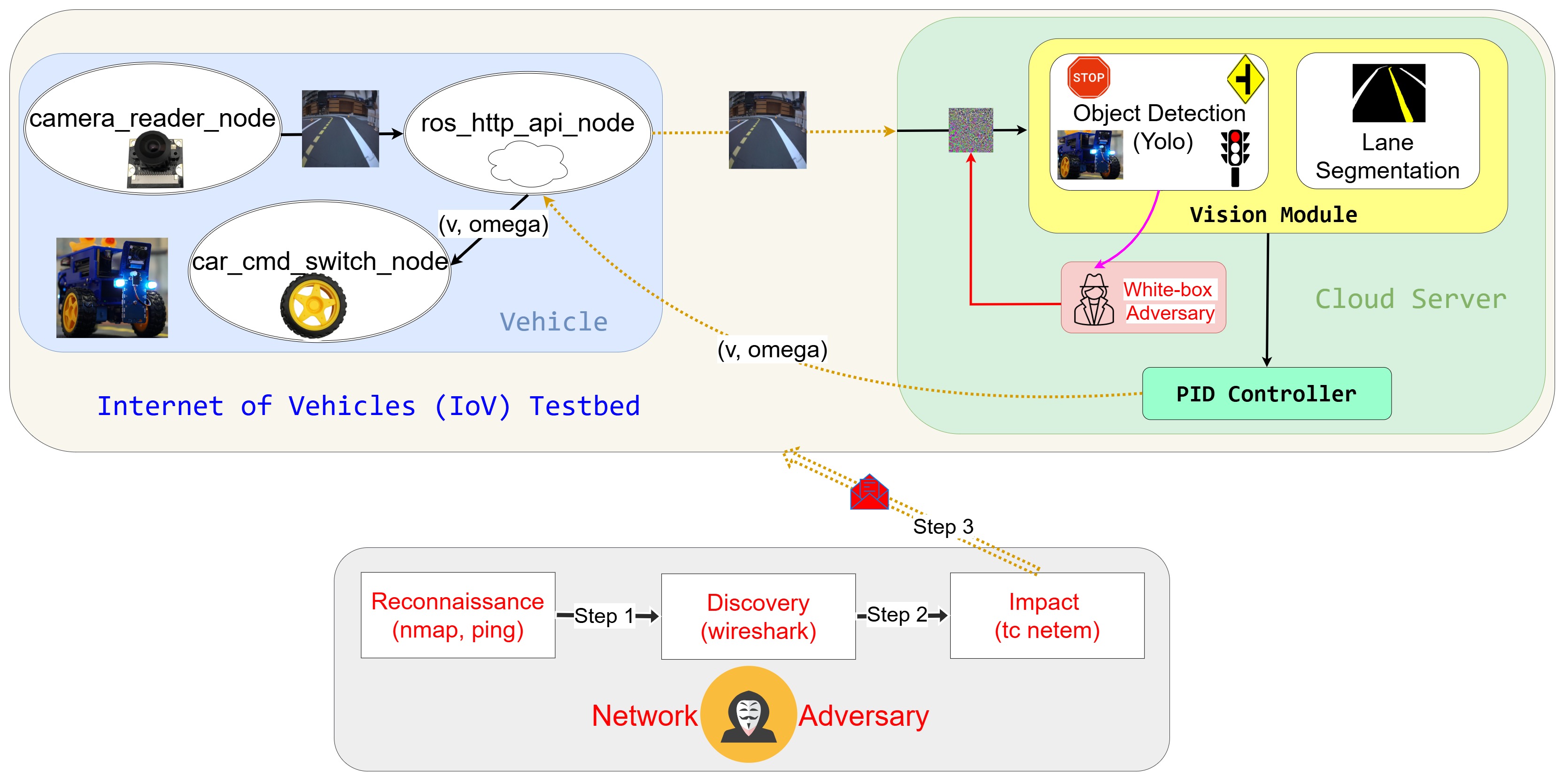}
    \caption{Architecture of the cloud-assisted autonomous driving testbed, illustrating the interaction between the vehicle, cloud vision–control loop, and adversarial agents.}
    \label{fig:architecture}
    \vspace{-3mm}
\end{figure*}

Figure~\ref{fig:architecture} shows the IoV testbed used to assess the robustness of cloud-assisted autonomous driving. The system comprises three subsystems: (i) a vehicle module for sensing and actuation, (ii) a cloud server executing perception and control, and (iii) an adversary module that executes white-box adversarial AI and network layer threats.

\subsection{Vehicle Module}
The vehicle functions as the edge client, operating on the Robot Operating System (ROS), a distributed, message-driven middleware widely adopted for robotic and autonomous systems. ROS provides a modular framework through its publisher--subscriber architecture, ensuring deterministic data exchange among sensing, decision, and actuation processes. The onboard camera (\texttt{camera\_reader\_node}) subscribes to the \texttt{/camera\_node/image/compressed} topic and transmits real-time frames to the cloud via a lightweight TCP bridge (\texttt{ros\_http\_api\_node}) over a persistent socket interface. Control commands generated by the cloud are routed back to the vehicle through the \texttt{car\_cmd\_switch\_node}, which interfaces directly with the wheel drivers to complete the perception–decision–action loop.

\subsection{Cloud Module}
\label{subsec:cloud_module}

The cloud server hosts the centralized perception and control pipeline, serving as the computational core of the Internet of Vehicles (IoV) architecture. Upon receiving image frames from the vehicle through a TCP-based ROS interface, the perception stack executes YOLOv8 for object detection (e.g., stop signs, traffic lights, vehicles) and U-Net for lane segmentation. The outputs, consisting of bounding boxes, segmentation masks, and lane deviation estimates, are passed to a Proportional–Integral–Derivative (PID) controller that generates velocity $(v)$ and angular velocity $(\omega)$ commands.


\begin{figure}[!h] 
\centering 

\includegraphics[width=\columnwidth]{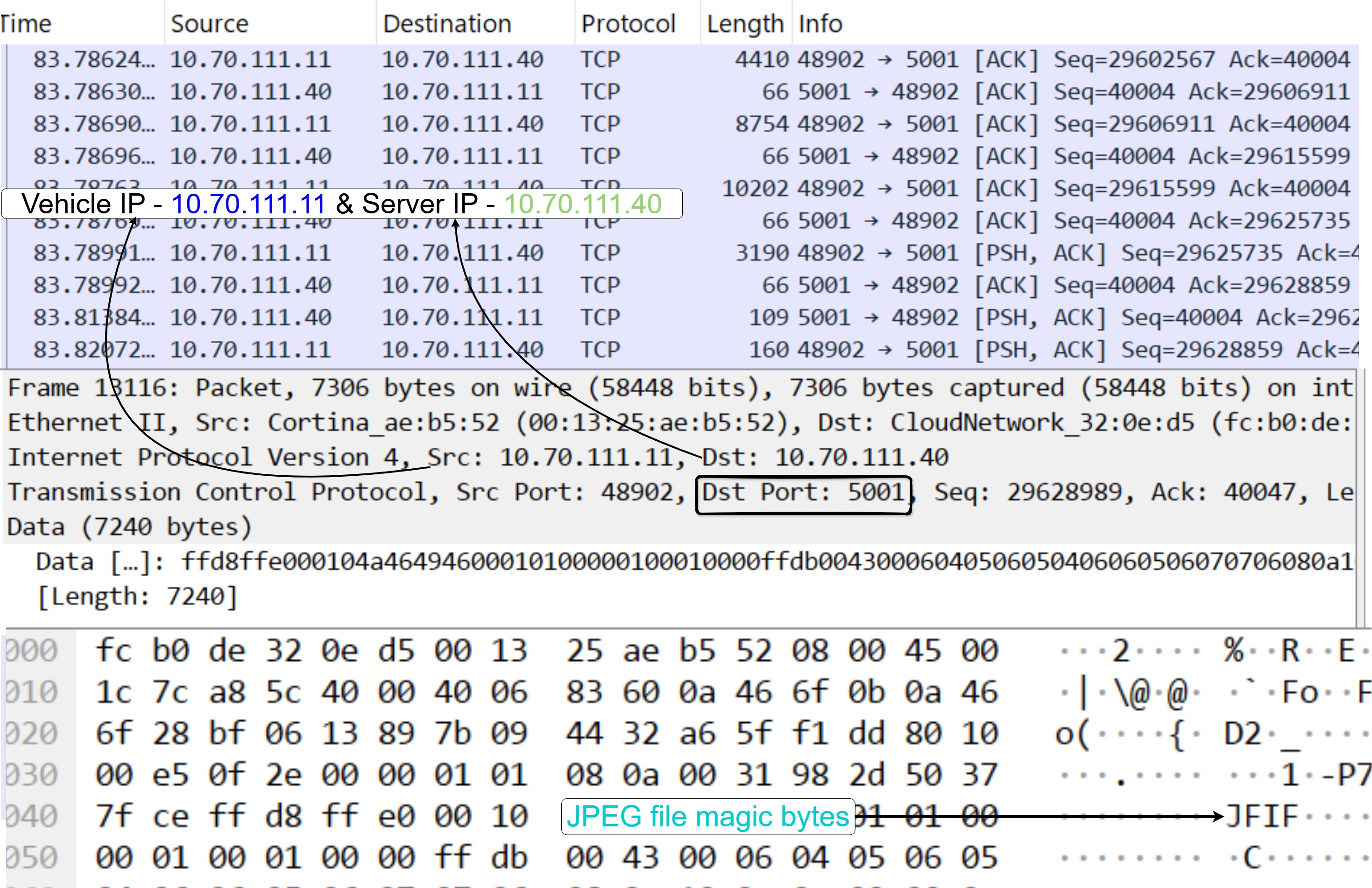} 
\caption{Wireshark capture showing TCP communication between the vehicle (10.70.111.11) and the cloud server (10.70.111.40) on port 5001. The highlighted frame reveals JPEG file transmission with identifiable magic bytes, confirming image payload exchange in the vehicle–cloud loop.} \label{fig:wireshark} 
\vspace{-5mm}
\end{figure}

\subsection{Adversary Module}
\label{subsec:adversary_module}

The Adversary Module captures both AI model-level and network-level threat actors simulated within the IoV testbed. It consists of: (i) \textit{Adversarial AI}, representing a white-box attacker embedded within the cloud pipeline, and (ii) \textit{Network Adversary}, representing an external agent targeting the vehicle–cloud commmunication.

\paragraph{Adversarial AI}
The adversarial AI is modeled as a white-box attacker with full access to the YOLOv8 detector, including its architecture, parameters, and training loss. This assumption enables direct gradient-based manipulation of input frames prior to inference. We implement FGSM and PGD using the supervised YOLOv8 loss in Eq.~\ref{eq:yolov8_loss} to generate adversarial perturbations. These perturbations are injected into the cloud-side perception pipeline to assess their impact on detection accuracy, class confidence, and downstream control behavior under adversarial conditions.

\paragraph{Network Adversary}
The network attacker follows a multi-stage sequence consistent with the MITRE ATT\&CK framework: (i) \textit{Reconnaissance}, where tools such as \texttt{nmap} and \texttt{ping} identify active hosts and open ports; (ii) \textit{Discovery}, where \texttt{Wireshark} is used to inspect ROS message streams and timing intervals; and (iii) \textit{Impact}, where \texttt{tc netem} injects controlled latency, jitter, and packet loss into the vehicle–cloud channel. These manipulations emulate network degradation of IoV systems.


\subsection{Workflow}
The testbed orchestrates synchronized operation across perception, control, and communication layers to emulate realistic cloud-assisted autonomy. The vehicle continuously streams image data to the cloud for inference and control computation as illustrated in Fig.~\ref{fig:wireshark} while adversarial modules perturb both perception and communication pipelines. The resulting closed-loop framework enables a reproducible real-time evaluation of system resilience under compound threat conditions. Specifically, it allows controlled experimentation on how visual perturbations and network attack affect perception reliability, control stability, and safety in cloud-assisted autonomous driving.

\begin{figure*}[t]
\centering
\includegraphics[
width=1\textwidth,
height=0.55\textheight,
keepaspectratio
]{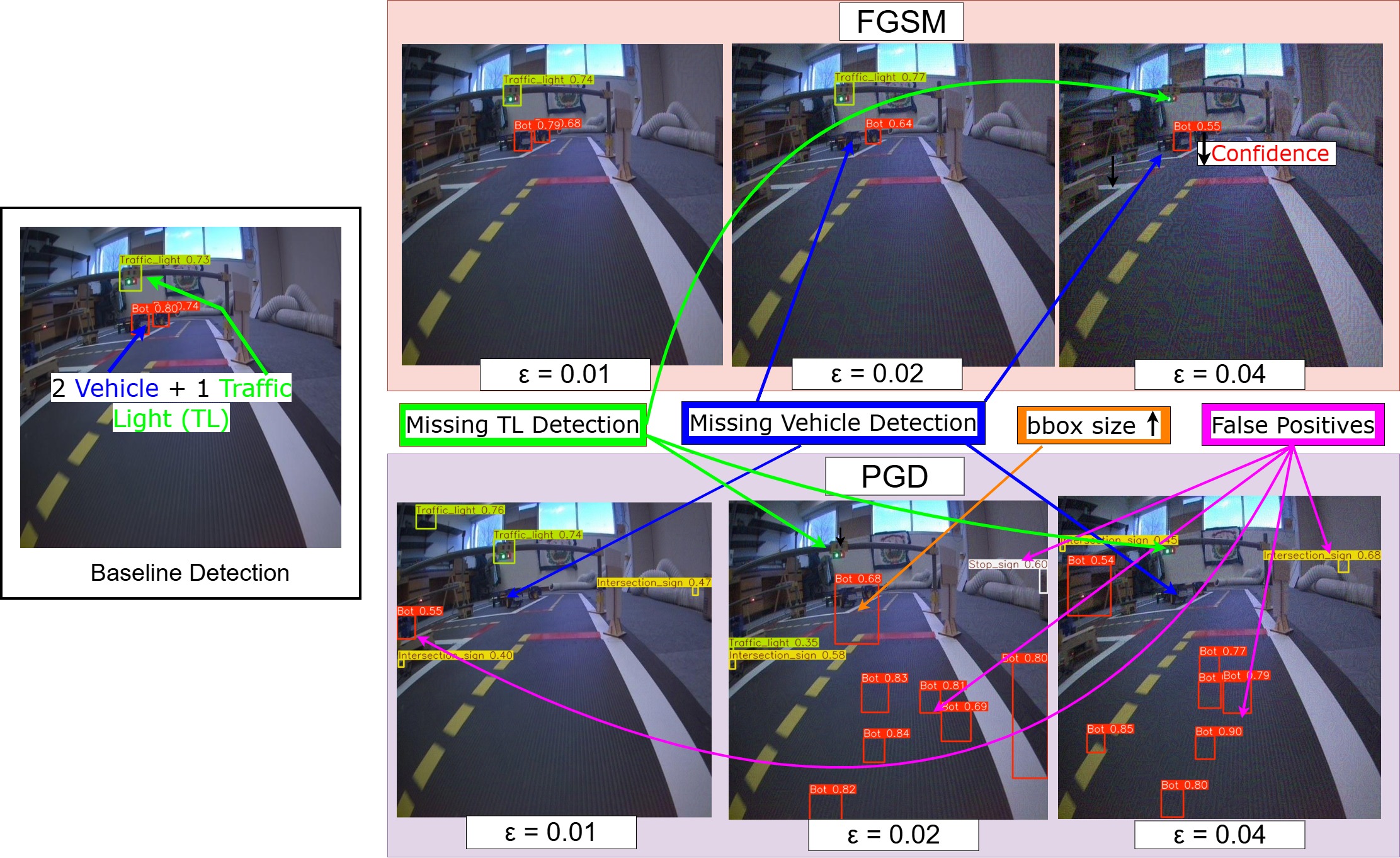}
\caption{The clean image serves as the baseline, followed by FGSM and PGD attacks with increasing perturbation magnitudes ($\epsilon$ = 0.01, 0.02, 0.04). Adversarial perturbations lead to missed vehicle and traffic light detections as well as false positives, demonstrating progressive degradation in detection reliability.}
\label{fig:adv_yolo}
\end{figure*}

\section{Experimental Setup}
\label{sec:exp_setup}

All experiments are conducted on the proposed IoV testbed, consisting of an on-vehicle edge client (Duckiebot with Jetson board) and a cloud perception--control server (Ubuntu 22.04 workstation, NVIDIA RTX A2000 12\,GB and 32\,GB RAM). Duckiebot streamed $640\times640$ frames to the server for inference; server-computed velocity and angular commands were returned to the vehicle and executed for closed-loop operation. The YOLOv8n model was trained on $2{,}414$ annotated Duckiebot RGB frames (70/30 - train/test split) with pixel-level lane masks and bounding boxes for traffic signs, lights, vehicles, and static obstacles.

FGSM and PGD are used as white-box adversarial attacks, applied per frame prior to YOLOv8 inference. FGSM generates a single-step perturbation from the supervised YOLOv8 loss, while PGD iteratively refines this perturbation under an $L_\infty$ constraint to produce stronger adversarial examples. Attack hyperparameters are set as follows:
\begin{itemize}
  \item \textbf{Perturbation budgets:} $\epsilon \in \{0.01,\,0.02,\,0.04\}$; corresponding to $\{2/255,\,4/255,\,8/255\}$ in pixel intensity.

  \item \textbf{PGD settings:} step size $\alpha = 0.01$, iterations $T = 10$.
\end{itemize}

\section{Results and Discussion}
\label{sec:results}
We begin with a qualitative assessment of adversarial effects on a representative test frame. Fig.~\ref{fig:adv_yolo} visualizes the evolution of YOLOv8 detections as the perturbation strength $\epsilon$ increases. In the clean image, the detector correctly identifies two vehicles and a traffic light with high confidence. Under FGSM, detection confidence progressively degrades, resulting in missed vehicle detections at higher $\epsilon$, while most bounding boxes remain spatially consistent. PGD produces a more disruptive effect: false positives appear at lower perturbation levels, and at larger $\epsilon$ multiple vehicles and traffic lights are entirely suppressed. These observations indicate that iterative attacks induce more structured and destructive failure modes than single-step perturbations, as each gradient update progressively steers the input toward regions of higher model loss. Despite the perturbations remaining visually imperceptible, the resulting perception errors demonstrate the sensitivity of deep detection models to adversarial manipulation.

\begin{figure}[!htp]
    \centering
    \includegraphics[width=\linewidth]{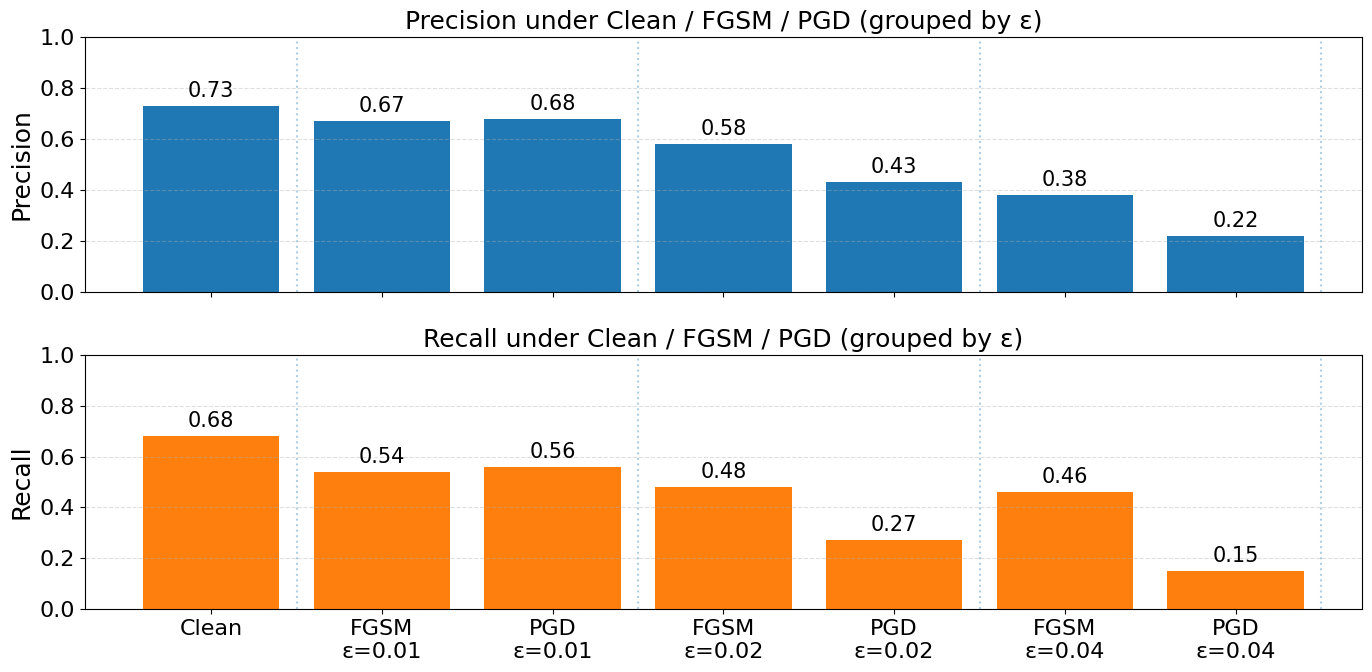}
    \caption{Precision–Recall comparison under clean and adversarial conditions (FGSM \& PGD) across $\epsilon = \{0.01, 0.02, 0.04\}$.}
    \label{fig:precision_recall}
\end{figure}

To quantify these effects on the full test set, Fig.~\ref{fig:precision_recall}  presents precision and recall under increasing perturbation strength. Both metrics decline steadily with $\epsilon$, indicating degraded object localization and classification. Under FGSM, precision drops from 0.73 to 0.38 and recall from 0.68 to 0.46 at $\epsilon=0.04$. PGD causes a substantially stronger degradation, reducing precision and recall to approximately 0.22 and 0.15, corresponding to nearly 70\% and 78\% reductions relative to the clean baseline. These results demonstrate that even small perturbation budgets can significantly compromise object detection reliability.

\begin{figure}[!htp]
    \centering
    \includegraphics[width=0.48\textwidth]{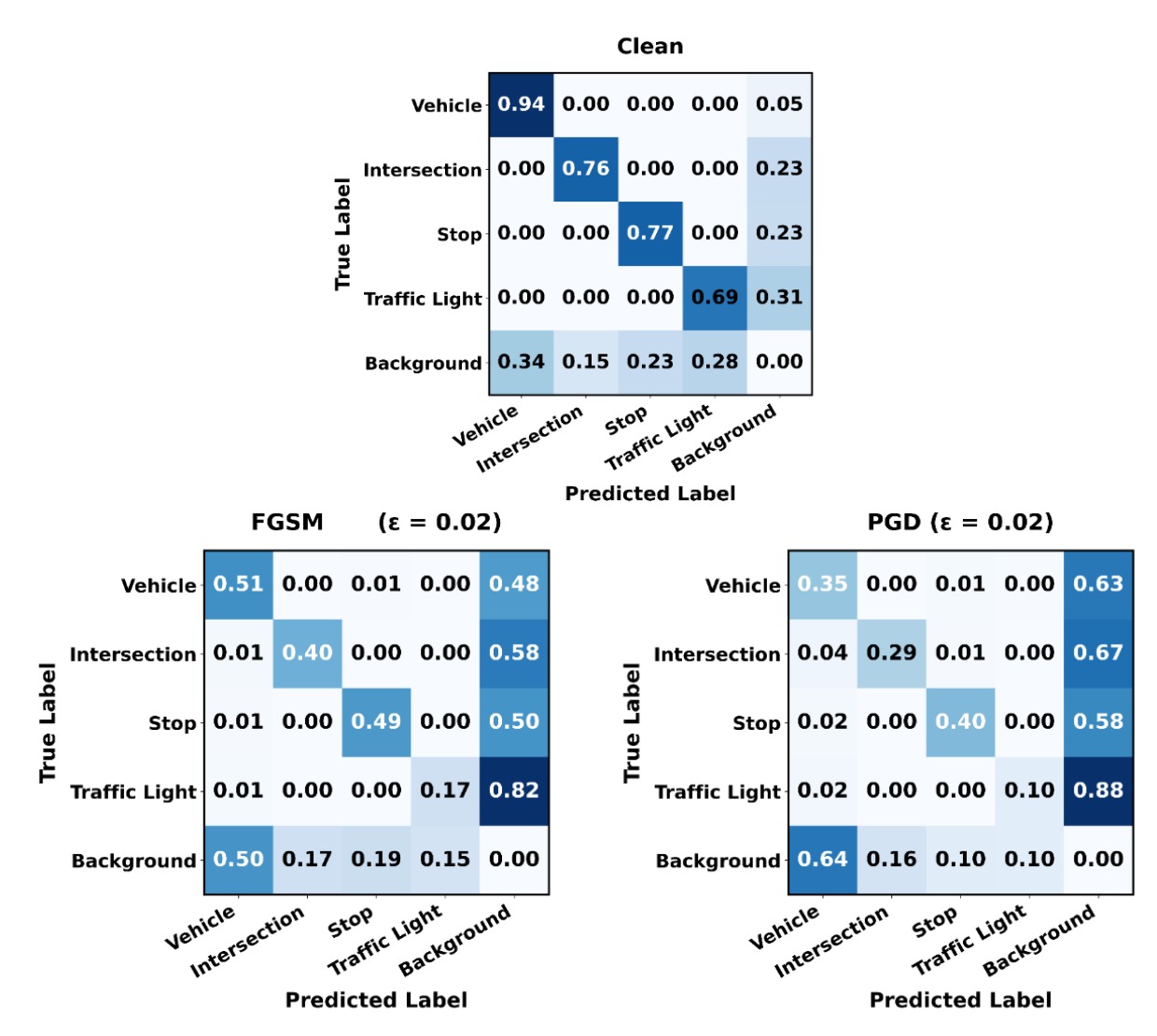}
    \caption{Confusion matrices for Clean, FGSM and PGD scenarios under $\epsilon = 0.02$.}
    \label{fig:confusion_matrices}
\end{figure}

We further analyze class-level behavior using confusion matrices at $\epsilon = 0.02$ (Fig.~\ref{fig:confusion_matrices}). The clean model exhibits strong diagonal dominance, indicating well-separated decision boundaries. Under FGSM, diagonal entries weaken as predictions shift toward the \emph{Background} class, reflecting reduced object confidence. PGD produces more severe distortion, with non-vehicle classes such as \emph{Stop}, \emph{Traffic Light}, and \emph{Intersection} frequently misclassified as \emph{Vehicle} or \emph{Background}. This behavior indicates that iterative perturbations disrupt semantic feature representations, leading to a breakdown of detection class separation.

\begin{figure}[!htbp]
    \centering
    \includegraphics[width=\linewidth]{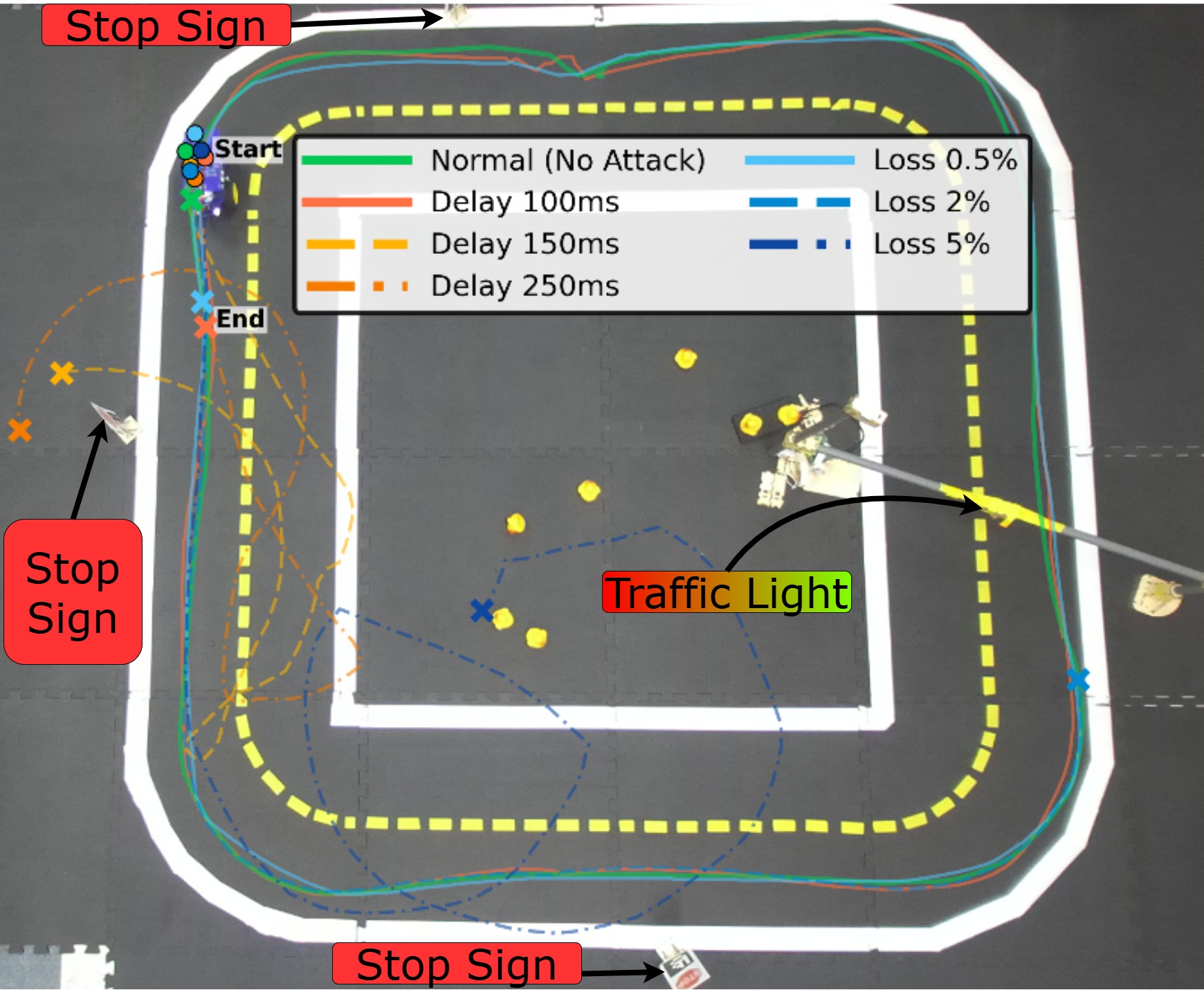}
    \caption{Vehicle trajectories under varying network delays and packet loss conditions on a rectangular track.}
    \label{fig:trajectory_overlay}
\end{figure}

We then evaluate the impact of network adversaries on closed-loop vehicle control in a structured urban track containing three stop signs and one traffic light. As shown in Fig.~\ref{fig:trajectory_overlay}, the vehicle maintains accurate lane tracking under nominal conditions. Trajectories under delay are shown in \textcolor{orange}{orange} and those under packet loss in \textcolor{blue}{blue}, with solid, dashed, and dash--dot lines representing increasing impairment severity (100~ms~$\rightarrow$~150~ms~$\rightarrow$~250~ms delay, and 0.5\%~$\rightarrow$~2\%~$\rightarrow$~5\% loss). The solid \textcolor{green}{green} line indicates the baseline trajectory without network attack. Moderate delay (100~ms) or minor packet loss (0.5\%) causes slight trajectory deviations near corners due to delayed actuation. At intermediate levels (150~ms delay or 2\% loss), trajectories exhibit lateral oscillations and reduced control stability, particularly at intersections. Under severe degradation (250~ms delay or 5\% loss), the control loop destabilizes as commands arrive too late or are intermittently dropped, resulting in lane departures and incomplete laps.

Table~\ref{rule_compliance} summarizes stop-sign compliance across different network conditions. Under nominal conditions and mild impairments (100~ms delay or 0.5\% packet loss), the vehicle successfully stops at all three intersections. However, as delay increases to 150~ms or packet loss rises to 2\%, violations begin to appear, particularly at downstream intersections where delayed perception updates accumulate. Under severe network degradation (250~ms delay or 5\% packet loss), the vehicle fails to stop at multiple intersections.

\begin{table}[!htb]
\centering
\renewcommand{\arraystretch}{0.9}
\setlength{\tabcolsep}{7pt}
\begin{tabular}{
    >{\raggedright\arraybackslash}m{2.7cm}
    >{\centering\arraybackslash}m{1.3cm}
    >{\centering\arraybackslash}m{1.3cm}
    >{\centering\arraybackslash}m{1.3cm}
}
\toprule
\textbf{Scenario} & \textbf{Stop 1} & \textbf{Stop 2} & \textbf{Stop 3} \\
\midrule
Normal (No Attack) & \textcolor{green!60!black}{\ding{51}} & \textcolor{green!60!black}{\ding{51}} & \textcolor{green!60!black}{\ding{51}} \\
Delay 100 ms       & \textcolor{green!60!black}{\ding{51}} & \textcolor{green!60!black}{\ding{51}} & \textcolor{green!60!black}{\ding{51}} \\
Delay 150 ms       & \textcolor{red!70!black}{\ding{55}}   & \textcolor{red!70!black}{\ding{55}}   & \textcolor{red!70!black}{\ding{55}}   \\
Delay 250 ms       & \textcolor{red!70!black}{\ding{55}}   & \textcolor{red!70!black}{\ding{55}}   & \textcolor{red!70!black}{\ding{55}}   \\
Loss 0.5\%         & \textcolor{green!60!black}{\ding{51}} & \textcolor{green!60!black}{\ding{51}} & \textcolor{red!70!black}{\ding{55}}   \\
Loss 2\%           & \textcolor{green!60!black}{\ding{51}} & \textcolor{red!70!black}{\ding{55}}   & \textcolor{red!70!black}{\ding{55}}   \\
Loss 5\%           & \textcolor{green!60!black}{\ding{51}} & \textcolor{red!70!black}{\ding{55}}   & \textcolor{red!70!black}{\ding{55}}   \\
\bottomrule
\end{tabular}
\caption{\textbf{Traffic Rule Compliance under Network Attack Scenarios.}}
\label{rule_compliance}
\end{table}

These results demonstrate that timing disruptions in the vehicle--cloud communication loop degrade the temporal consistency of control execution, even when perception outputs remain accurate. Consequently, network-layer attacks can lead to unsafe driving behavior, including missed traffic rules and unstable trajectory tracking. When combined with adversarial perception errors, such communication impairments further amplify instability in the perception--control loop, highlighting a critical cross-layer vulnerability in cloud-assisted autonomous driving systems.

\section{Conclusion}
\label{sec:conclusion}

This work presents a hardware-in-the-loop Internet of Vehicles (IoV) testbed to evaluate the robustness of cloud-assisted autonomous driving under coordinated adversarial conditions. The framework integrates real-time perception, control, and communication, enabling systematic analysis of how adversarial perturbations and network-level attacks influence closed-loop control stability. Experimental results show that visually imperceptible image perturbations at $\epsilon = 0.04$ significantly reduce detection precision and recall, while network impairments in the range of 100–250\,ms delay or 0.5–5\% packet loss, corresponding to transient losses of approximately three to four frames, introduce substantial latency and synchronization errors. When combined, these degradations amplify instability in perception–control coupling, leading to delayed actuation, trajectory drift, missed stops, and traffic rule violations. These findings underscore the importance of cross-layer resilience in cloud-assisted autonomy, supported by adversarially robust perception models, latency-aware control strategies, and adaptive communication mechanisms that can tolerate transient network disruptions.

\section{Acknowledgment}
The Authors would like to thank DARPA AI-CRAFT AWD16069 grant for the partial financial support for this work.

\bibliographystyle{IEEEtran}
\bibliography{bibliography}

\end{document}